\documentclass{article}
\pdfoutput=1
\usepackage{spconf,amsmath,graphicx}
\usepackage{amsfonts}
\usepackage{booktabs,adjustbox}
\usepackage[dvipsnames]{xcolor}
\usepackage{hyperref}
\hypersetup{
	colorlinks,
	linkcolor={red!50!black},
	citecolor=ForestGreen,
	urlcolor={blue!80!black}
}
\newcommand{\etal}{\textit{et al}.}
\newcommand{\ie}{\textit{i}.\textit{e}.}

\newcommand{\etc}{\textit{etc}.}
\def\dashdotted{\xleaders\hbox to 1em{$-\cdot$}\hfill$-$}

\title{Self-supervised Representation Learning for Ultrasound Video}
\name{Jianbo Jiao$^1$, Richard Droste$^1$, Lior Drukker$^2$, Aris T. Papageorghiou$^2$, J. Alison Noble$^1$\thanks{We acknowledge the EPSRC (EP/M013774/1, Project Seebibyte), ERC (ERC-ADG-2015 694581, Project
PULSE), and the support of NVIDIA Corporation with the donation of the Titan V GPU.}}
\address{$^1$Institute of Biomedical Engineering, Department of Engineering Science, University of Oxford, UK\\
$^2$Nuffield Department of Women’s \& Reproductive Health, University of Oxford, UK}
\begin{document}
%\ninept
%
\maketitle
\begin{abstract}
  Recent advances in deep learning have achieved promising performance for medical image analysis, while in most cases ground-truth annotations from human experts are necessary to train the deep model. In practice, such annotations are expensive to collect and can be scarce for medical imaging applications. Therefore, there is significant interest in learning representations from unlabelled raw data. In this paper, we propose a self-supervised learning approach to learn meaningful and transferable representations from medical imaging video without any type of human annotation. We assume that in order to learn such a representation, the model should identify anatomical structures from the unlabelled data. Therefore we force the model to address anatomy-aware tasks with free supervision from the data itself. Specifically, the model is designed to correct the order of a reshuffled video clip and at the same time predict the geometric transformation applied to the video clip. Experiments on fetal ultrasound video show that the proposed approach can effectively learn meaningful and strong representations, which transfer well to downstream tasks like standard plane detection and saliency prediction.
\end{abstract}
\begin{keywords}
Self-supervised, representation learning, ultrasound video
\end{keywords}
\section{Introduction}
\label{sec:intro}
Machine learning, especially deep learning techniques, have witnessed great success in medical image analysis in recent years.
Several learning-based approaches have demonstrated superior performance over human experts~\cite{ardila2019end,hannun2019cardiologist}.
Most of these methods extensively rely on ground-truth labeled data annotated by human experts, to train the deep model.
However, data annotation is expensive to scale, and in addition, the ``ground-truth'' label of medical images might be inaccessible.
Therefore, in this paper, we are interested in the question: \emph{Is it possible to learn meaningful representations directly from raw data, without any human annotations?}

Here we explore this question through self-supervised representation learning, in which ``self-supervised'' indicates that the learning process is supervised purely based on the data itself (also termed unsupervised in some literature).
In our work, we define representation learning as capturing anatomy-aware knowledge, which plays a crucial role in medical image interpretation.
Specifically, we showcase the effectiveness of the proposed representation learning approach with application to automated fetal ultrasound (US) scan interpretation.

Visual representation learning has been recently explored for natural images. For instance, Zhang \etal~\cite{zhang2016colorful} propose to learn representations by colourising a grayscale image to its color version equivalent.
However, such a colourisation strategy does not apply to US images (and other medical images) due to its monotone nature and absence of true colour.
Wang and Gupta~\cite{wang2015unsupervised} learn visual representations from video data by using off-the-shelf tracking algorithms.
Whereas its learning ability is restricted by the performance of the used tracker and visual tracking in US is a difficult task itself.
On the other hand, for US and even medical images in general, self-supervised representation learning is under-explored.
Given that medical annotation is expensive and infeasible in many cases, it is crucial to learn representations from medical images without annotation.

In the field of fetal US image analysis, recent works on standard plane detection and visual saliency prediction show promising performance with deep learning.
Baumgartner \etal~\cite{baumgartner2017sononet} proposed a convolutional neural network (CNN) to detect 2D fetal standard views.
Cai \etal~\cite{cai2018multi} augment the standard plane detection with a visual saliency prediction task and show that it helps the detection task.
Inspired by the saliency prediction task, a recent work~\cite{droste2019ultrasound} shows that transferable representations can be learned by modelling sonographer visual attention.
To our knowledge, this is the only US representation learning method in the literature.
However, it fully relies on gaze-tracking data as a ground-truth label for supervision, limiting its generalisability as an approach.

Different from the abovementioned prior works, in this paper we propose a new self-supervised representation learning approach tailored for characteristics of fetal US video.
Specifically, based on the assumption that distinctive strong representations should be anatomy-aware for US data, we design a joint reasoning task to force the model to identify anatomical structures during the self-supervised learning process: correct the reshuffled video frames and identify the specific transformation applied to the video clip simultaneously.
To the best of our knowledge, this is the first attempt to learn representations from US video without any type of external annotation.
The learned representations are evaluated on two US tasks -- standard plane detection and saliency prediction.
The weights learned by the proposed approach are fine-tuned on the above two tasks with limited training samples, to demonstrate the transferability of our method.
Extensive experimental results show that the proposed self-supervised learning approach is able to capture meaningful and strong representations that transfer well to downstream tasks, with a small margin towards the performance of supervised methods.

The main contributions of our work are: 1) We present, to our knowledge, the first attempt towards self-supervised representation learning for fetal US video, without any external annotations; 2) We propose a joint reasoning approach to implicitly force the model to learn anatomy-aware representations; 3) Even with only the raw data itself, our approach is demonstrated to be effective in learning transferable representations, by evaluations on various downstream tasks.

\section{Methods}
\label{sec:method}
In this section, we describe the proposed self-supervised representation learning approach in detail.
The main idea is that if the learned representation is strong and with good transferability, the corresponding deep model should be capable of identifying the anatomical structures in the data.
In this paper, we demonstrate the idea with fetal US video data.
The specially designed tasks are described as follows.

\subsection{Temporal Order Correction}
\begin{figure}
  \centering
  \includegraphics[width=0.9\columnwidth]{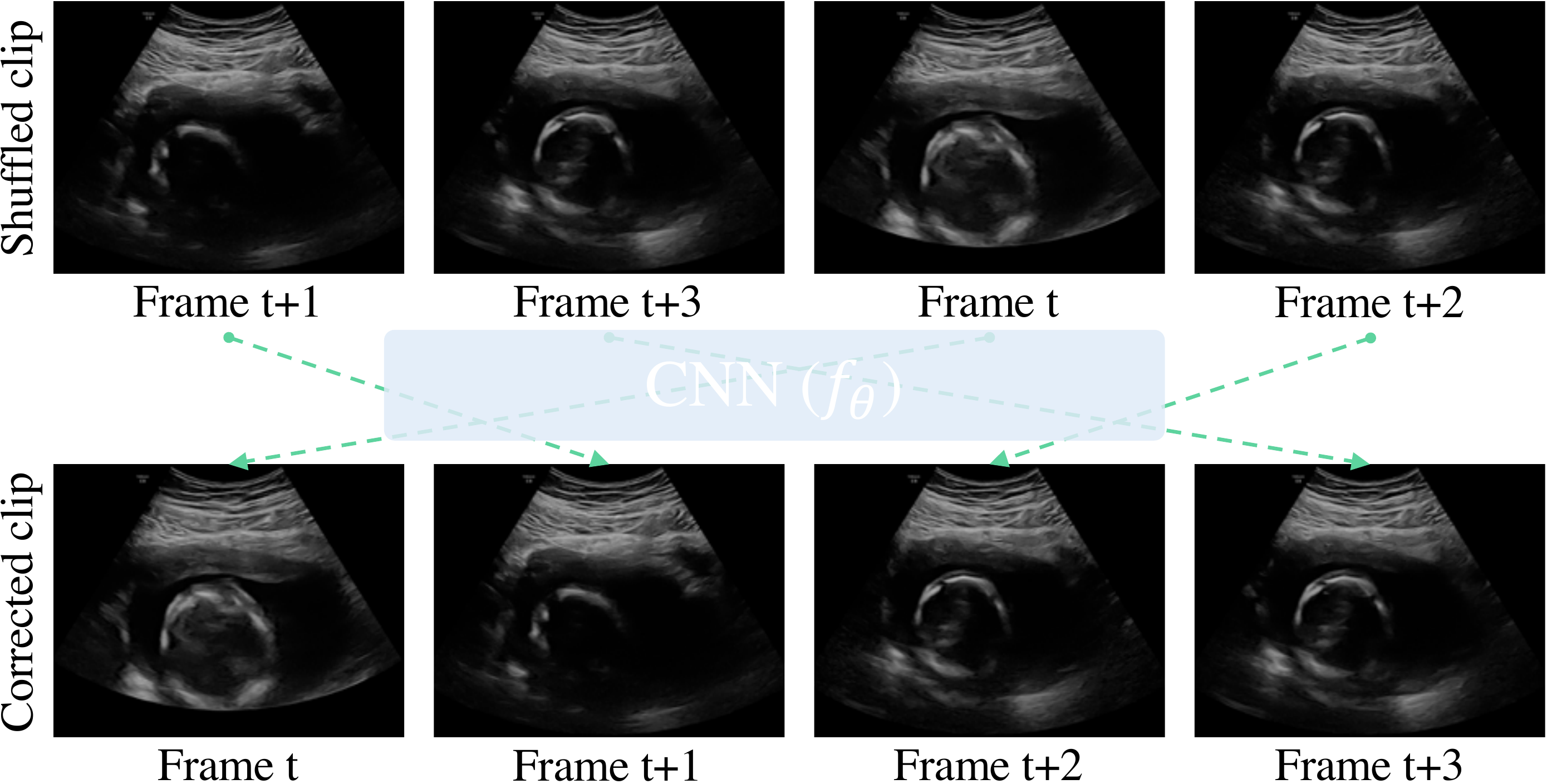}
  \caption{Illustration of the temporal order correction. Randomly shuffled video frames are fed into the CNN model $f_{\theta}$ to correct the frame order.}
  \label{fig:ord}
\end{figure}

Suppose $x_i\in \mathcal{X}\subset\mathbb{R}^{H\times W\times K}$ is a video frame in the video data set $\mathcal{X}$, where $H, W$ and $K$ represent the height, width and number of frames in a video clip $\{x_1, x_2,..., x_K\}$.
Here we first explore representation learning by designing the task of temporal order correction.
We argue that if a model can correct a randomly shuffled video clip, it should be aware of the underlying anatomical representations.
A video clip is pre-processed by reshuffling the frame order and taken as input to a CNN model $f_{\theta}$.
The model is trained to correct the random order and reconstruct the original video clip.
An example is illustrated in Fig.~\ref{fig:ord}.
Taking $K=4$ as an example, the index of the reshuffled video clip \{1,3,0,2\} (or \{2,0,3,1\}) is corrected by the model to \{0,1,2,3\}.
To capture high-level context information, here we model the order correction task as a classification problem.
Specifically, the model takes a reshuffled video clip as input and predict the correct order index of this clip.
For $K=4$ the correction space is with size of $^4P_4/2=12$, \ie, a 12-way classification problem.

\subsection{Spatio-temporal Transform Prediction}
\begin{figure}
  \centering
  \includegraphics[width=0.9\columnwidth]{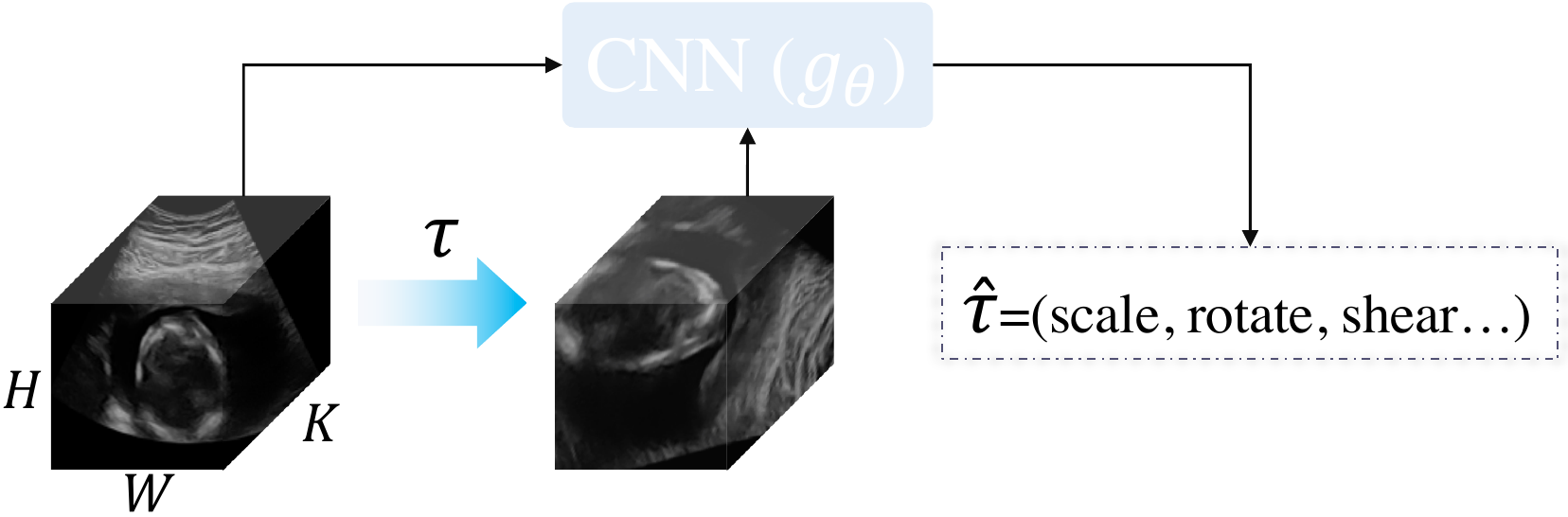}
  \caption{Illustration of the spatio-temporal transform prediction. The model $g_{\theta}$ is learned to predict the transformation $\tau$ applied to the video clip.}
  \label{fig:trans}
\end{figure}

In addition to the temporal representation, we also take spatial information into consideration.
Here the spatial representation is learned through a task of predicting the parameters of the affine transformation (\ie, translation, scale, rotation, and shear) that is applied to a video clip.
We design this task based on the assumption that the model should identify distinctive structures in order to correctly predict the specific transformation.
Given a video clip $v=\{x_1,x_2,...,x_K\}$ and a spatio-temporal transformation $\tau\in\mathcal{T}$ where $\mathcal{T}$ is a transformation set, the video clip is transformed by $\tau(v)$ and sent to a CNN model $g_{\theta}$ together with the original clip $v$.
Then the model is trained to predict the applied transformation $\tau$.
Although various transformations can be included into set $\mathcal{T}$, for simplicity we use the affine transform in this paper.
An example is illustrated in Fig.~\ref{fig:trans}.
Affine transformations of scale, rotate, shear, \etc~are applied to the original video clip.
Based on sufficient training data and random transformations, the model $g_\theta$ learns to predict the transformation parameters accordingly.

\subsection{Joint Anatomy-Aware Reasoning}
\begin{figure}
  \centering
  \includegraphics[width=\columnwidth]{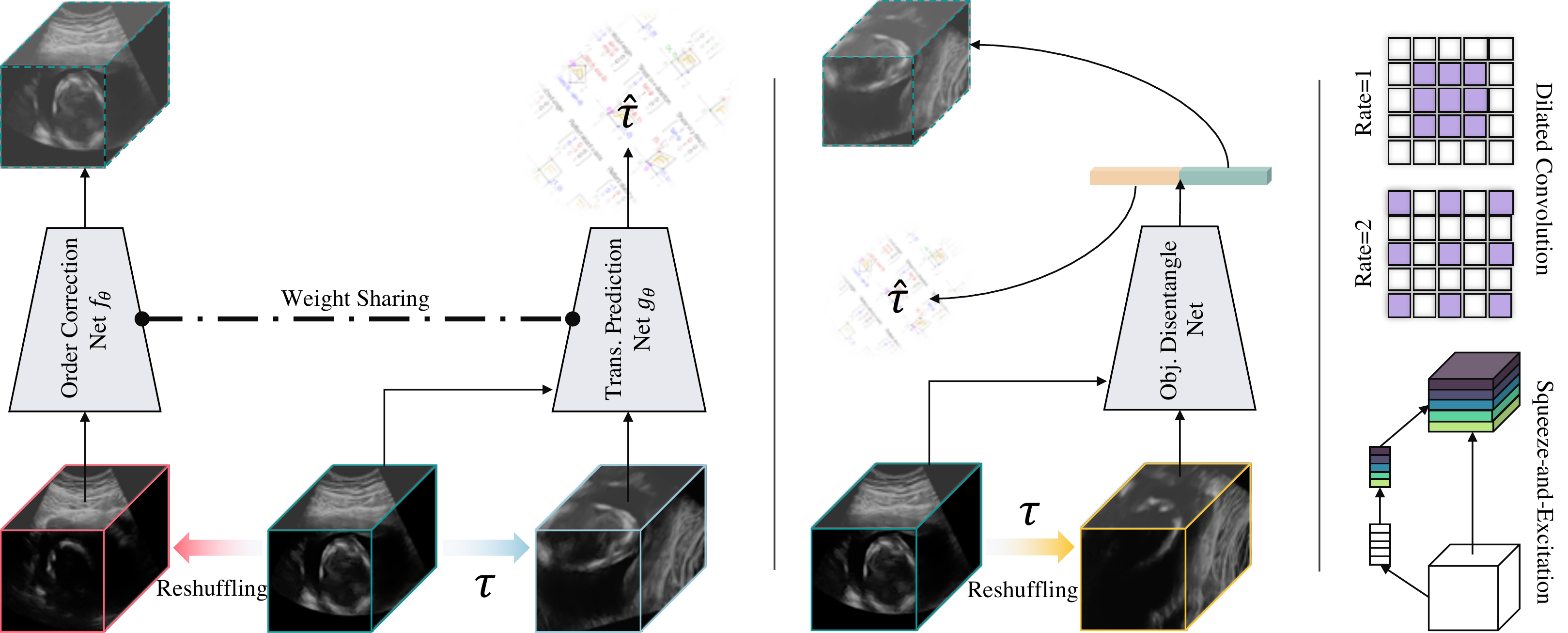}
  \caption{The proposed joint reasoning frameworks. Left to right: partially Siamese network;  objective disentangle network; More details~\cite{hu2018squeeze} for the network components.}
  \label{fig:net}
\end{figure}

Based on the abovementioned tasks, we propose a joint reasoning approach that leverages the learning ability from both order correction and geometric transformation prediction for anatomy-aware representation learning.
Here we explore two strategies to combine these tasks: a partially Siamese network, or disentangling a joint objective.
Specifically, the first strategy is performing the two tasks in parallel while using a Siamese network that partially shares weights; the latter solution is to apply both the frame reshuffling and transformation onto the same video clip, and let a single model predict and disentangle these two features.
The network architectures of these solutions are shown in Fig.~\ref{fig:net}.
To train the joint reasoning networks, we define the objective function as follows:
\begin{equation}
  \mathcal{L}=\mathcal{L}_{ord}+\mathcal{L}_{trans},
\end{equation}
where $\mathcal{L}_{ord}$ is a cross-entropy loss for the index classification and $\mathcal{L}_{trans}=\|\hat\tau-\tau\|$ is for the transformation parameter prediction. Note that all the above supervisions are free to use, without any external annotations.

\subsection{Network Implementation}
We use the SENet~\cite{hu2018squeeze} as our backbone network, though other CNNs can also be easily applied.
Following prior works~\cite{droste2019ultrasound}, the normal convolutions are replaced with dilated convolutions, as illustrated in Fig.~\ref{fig:net} (right).
The models were trained with SGD optimizer with momentum of 0.9 and weight decay of $10^{-4}$.
Learning rate was set as 0.1 and the models were trained for 8 epochs.
Batch size is set as 32 and data augmentation was performed including random crop, horizontal flipping, and gamma and brightness variation.
The models were implemented in PyTorch on an NVIDIA Titan V GPU.

\section{Experiments and Results}
\label{sec:exp}
In our experiment, we used a routine clinical fetal US dataset\footnote{UK Research Ethics Committee Reference 18/WS/0051.} with both video sequences and corresponding real-time gaze-tracking data from sonographers.
The average video duration for each scan is about 82,000 frames.
In our experiments, 135 scans are used with three-fold cross-validation by equally-divided subsets (90 training and 45 testing).
By discarding invalid data and temporally down-sampling at a rate of eight, we acquired 400,000 frames in total.
The fan-shape was removed by centre-cropping to avoid trivial solution for our designed tasks and each video clip consists of four frames.

\subsection{Evaluation on Standard Plane Detection}
\label{sec:SPD}
\begin{figure}
  \centering
  \includegraphics[width=\columnwidth]{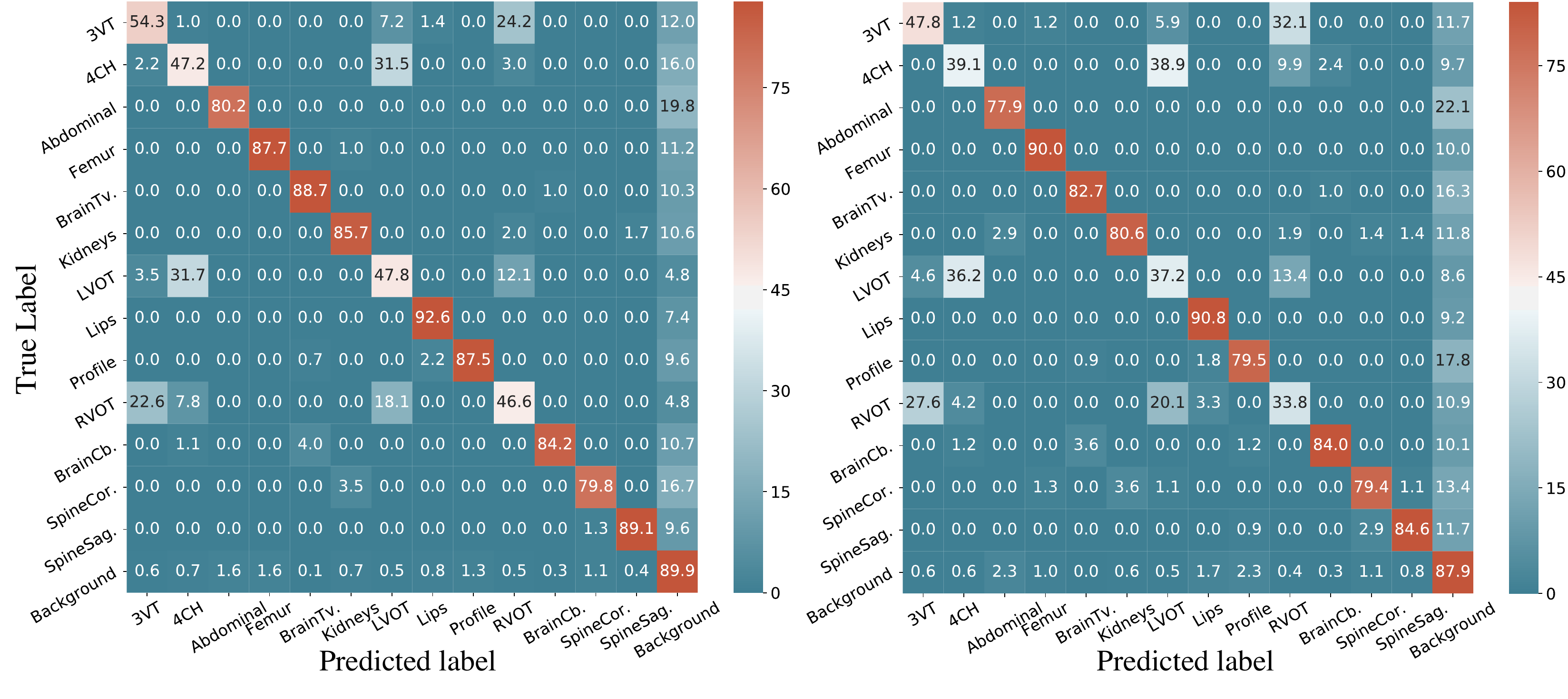}
  \caption{Confusion matrix of the precision on standard plane detection for \emph{Ours-Siam.} (left) and \emph{Ours-Dise.} (right).}
  \label{fig:cm}
\end{figure}

We first evaluate the effectiveness of our learned representations by transferring to the standard plane detection task.
Similar to~\cite{baumgartner2017sononet,droste2019ultrasound}, we consider 14 categories: three-vessel and trachea view (3VT), four-chamber view (4CH), left ventricular outflow tract (LVOT), right ventricular outflow tract (RVOT), profile, abdominal, brain cerebellum plane (BrainCb.), brain transventricular plane (BrainTv.), femur, kidneys, lips, spine coronal plane (SpineCor.), spine sagittal plane (SpineSag.) and background.
The learned weights from our models are loaded to a new model (same architecture except for the final classification layers) to fine-tune standard plane detection.
Baseline methods of random initialisation, initialisation from gaze/saliency prediction~\cite{droste2019ultrasound} and SonoNet~\cite{baumgartner2017sononet} were included for a comparison. The training settings were kept the same as~\cite{droste2019ultrasound} for fair comparison.
Evaluation results are shown in Table~\ref{tab:spd}.
\begin{table}
  \caption{Evaluation results on standard plane detection (mean$\pm$std.[\%]). Best performance is marked in \textbf{bold}. Note the last three methods use external labels/supervision.}
  \adjustbox{max width=\columnwidth}{
  \centering
  \begin{tabular}{@{}l|ccc|ccc@{}}
    \toprule
    & Rand.Init. & Ours-Siam. & Ours-Dise. & Gaze & Saliency & SonoNet \\
    \midrule
    Precision & 70.4{\small$\pm$2.3} & \textbf{75.8}{\small$\pm$1.9} & 71.1{\small$\pm$3.1} & \textcolor{gray}{ 67.2{\small$\pm$3.4}} & \textcolor{gray}{79.5{\small$\pm$1.7}} & \textcolor{gray}{82.3{\small$\pm$1.3}} \\
    Recall & 64.9{\small$\pm$1.6} & \textbf{76.4}{\small$\pm$2.7} & 71.9{\small$\pm$1.4} & \textcolor{gray}{57.3{\small$\pm$4.5}} & \textcolor{gray}{75.1{\small$\pm$3.4}} & \textcolor{gray}{87.3{\small$\pm$1.1}} \\
    F1-score & 67.0{\small$\pm$1.3} & \textbf{75.7}{\small$\pm$2.0} & 71.0{\small$\pm$2.3} & \textcolor{gray}{60.7{\small$\pm$3.9}} & \textcolor{gray}{76.6{\small$\pm$2.6}} & \textcolor{gray}{84.5{\small$\pm$0.9}} \\
    \bottomrule
  \end{tabular}
  }
  \label{tab:spd}
\end{table}
From the results we can observe that our approaches perform consistently better than the random initialisation and even better than the gaze-prediction method which leverages gaze as external supervision.
Our Siamese approach is slightly better than the disentangle one, which suggests that earlier isolation is more suitable for joint reasoning.
Moreover, our Siamese approach is more stable with a substantially lower standard deviation.
To better understand the contribution of each category, we present the confusion matrix for the precision of our two solutions in Fig.~\ref{fig:cm}.
We can see that for most cases our two approaches can correctly detect the standard plane, while performing relatively worse on the cardiac views (\emph{3VT, 4CH, LVOT, RVOT}).
We attribute this to the confusing representations among these categories, as even human experts are poor at distinguishing them.

\subsection{Evaluation on Saliency Prediction}
\begin{figure}
  \centering
  \includegraphics[width=\columnwidth]{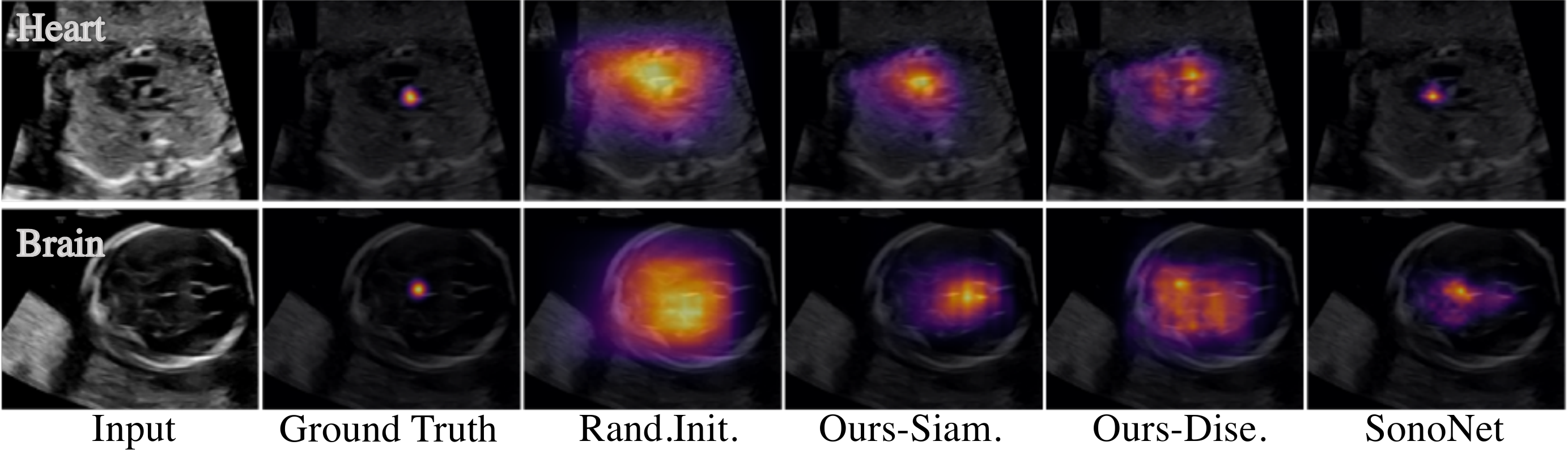}
  \caption{Qualitative performance for visual saliency prediction with comparison to alternative solutions and ground-truths.}
  \label{fig:sal}
\end{figure}

In addition to the classification-based task explored in Sec.~\ref{sec:SPD}, we evaluate the learned representation by a regression-based task, \ie, visual saliency prediction.
Similar to the evaluation on standard plane detection, we loaded the network weights that were learned by our approach to a saliency prediction model and fine-tune it.
We use a similar network structure with the final layers dilated to keep a spatial map for the regression task.
Since the representation learning methods proposed in~\cite{droste2019ultrasound} take gaze/saliency prediction as their training task, it is improper to include for comparison.
Instead, we compared with the random initialisation and initialise from the SonoNet weights.
The comparison result is shown in Table~\ref{tab:sal}, with evaluation metrics that are commonly used in saliency prediction works~\cite{bylinskii2018different,droste2019ultrasound}: Kullback-Leibler divergence (KL), normalised scanpath saliency (NSS), area under ROC curve (AUC), Pearson's correlation coefficient (CC) and similarity (SIM).
\begin{table}
  \caption{Evaluation results on visual saliency prediction. Best performance is marked in \textbf{bold}.}
  \centering
  \adjustbox{max width=\columnwidth}{
  \begin{tabular}{@{}l|ccccc@{}}
    \toprule
    & KL$\downarrow$ & NSS$\uparrow$ & AUC$\uparrow$ & CC$\uparrow$ & SIM$\uparrow$\\
    \midrule
    Rand.Init. & 3.94{\small$\pm$0.18} & 1.47{\small$\pm$0.24} & 0.90{\small$\pm$0.01} & 0.12{\small$\pm$0.02} & 0.05{\small$\pm$0.01}\\
    Ours-Siam. & \textbf{3.00}{\small$\pm$0.08} & \textbf{2.72}{\small$\pm$0.12} & \textbf{0.95}{\small$\pm$0.00} & \textbf{0.22}{\small$\pm$0.01} & \textbf{0.12}{\small$\pm$0.01} \\
    Ours-Dise. & 3.06{\small$\pm$0.08} & 2.54{\small$\pm$0.11} & 0.94{\small$\pm$0.00} & 0.21{\small$\pm$0.01} & 0.11{\small$\pm$0.01} \\
    \midrule
    SonoNet & \textcolor{gray}{3.14{\small$\pm$0.02}} & \textcolor{gray}{2.62{\small$\pm$0.03}} & \textcolor{gray}{0.94{\small$\pm$0.00}} & \textcolor{gray}{0.21{\small$\pm$0.00}} & \textcolor{gray}{0.12{\small$\pm$0.00}} \\
    \bottomrule
  \end{tabular}
  }
  \label{tab:sal}
\end{table}
The evaluation results in Table~\ref{tab:sal} again validate the effectiveness of the proposed approach.
Note that for this dense prediction task, the model with \emph{Ours-Siam}.\ weights performs even better than that from \emph{SonoNet} for some metrics.
Example qualitative performance is also presented for comparison in Fig.~\ref{fig:sal}, where consistent performance is shown with the quantitative results.

\subsection{Ablation Study}
For a better understanding of the contribution from each sub-task (temporal order correction and spatio-temporal transform prediction), we performed an ablation study and report the results in Table~\ref{tab:abl}.
\begin{table}
  \caption{Ablation study on the tasks of both standard plane detection and visual saliency prediction.}
  \centering
  \adjustbox{max width=\columnwidth}{
  \begin{tabular}{@{}l|ccc|ccccc@{}}
    \toprule
    & Prec.[\%] & Rec.[\%] & F1[\%] & KL$\downarrow$ & NSS$\uparrow$ & AUC$\uparrow$ & CC$\uparrow$ & SIM$\uparrow$\\
    \midrule
    Ord.Crct. & 72.1{\small$\pm$4.2} & 72.5{\small$\pm$2.1} & 71.8{\small$\pm$3.0} & 3.29{\small$\pm$0.14} & 2.29{\small$\pm$0.16} & 0.93{\small$\pm$0.01} & 0.19{\small$\pm$0.01} & 0.09{\small$\pm$0.01} \\
    Trans.Pred. & 72.8{\small$\pm$3.5} & 72.3{\small$\pm$1.7} & 72.3{\small$\pm$2.6} & 3.54{\small$\pm$0.61} & 1.97{\small$\pm$0.74} & 0.93{\small$\pm$0.02} & 0.16{\small$\pm$0.06} & 0.09{\small$\pm$0.02} \\
    Ours (final) & 75.8{\small$\pm$1.9} & 76.4{\small$\pm$2.7} & 75.7{\small$\pm$2.0} & 3.00{\small$\pm$0.08} & 2.72{\small$\pm$0.12} & 0.95{\small$\pm$0.00} & 0.22{\small$\pm$0.01} & 0.12{\small$\pm$0.01} \\
    \bottomrule
  \end{tabular}
  }
  \label{tab:abl}
\end{table}
It can be observed that either the order correction task or the transform prediction task performs quite well alone.
When jointly reasoning these two tasks in a Siamese model, the performance is further boosted.

\section{Discussion and Conclusion}
\label{sec:conc}
In this paper, for the first time, we addressed the self-supervised representation learning problem for fetal ultrasound video.
Transferable representations were learned without any type of external labels.
We assume meaningful and strong representations rely on the identification of anatomical information from the raw data, and proposed a joint reasoning framework to achieve that accordingly.
Extensive experiments on two US-related tasks show that the representations learned by our approach are meaningful and transferable, outperforming other alternative solutions.
Although in this work we showcase the self-supervised learning capability of the proposed approach on US, our framework is generic and has potential to be applied in other medical modalities, which we believe to be a promising direction for future work.

% References should be produced using the BibTeX program from suitable
\bibliographystyle{IEEEbib}
% \bibliography{refs}
% \small
\bibliography{refs_short}

\end{document}